\documentclass[10pt,conference]{ieeeconf} 
\IEEEoverridecommandlockouts                              % This command is only
\usepackage{cite}
\usepackage{float}
\usepackage{amsmath}
\usepackage{amsfonts}
\usepackage{graphicx}

\usepackage{booktabs}
\usepackage{placeins}
\usepackage{xr}
\usepackage{bbding}
\usepackage{array}

\usepackage{relsize}
\usepackage{graphicx}
\usepackage{esint}
\usepackage[normalem]{ulem}
\usepackage{arydshln}
\usepackage{eurosym}
\usepackage{color}
\usepackage{amsmath,amssymb}
\usepackage{multirow}
\usepackage{textcomp}
\usepackage{tabularx}
\usepackage{placeins}
\usepackage{lscape}
\usepackage{comment}
\usepackage{mathrsfs}
\usepackage{algorithm2e}
\usepackage{hyperref}       % hyperlinks
\usepackage[dvipsnames]{xcolor}
\usepackage{colortbl}
\usepackage{graphicx}

\usepackage{amsmath}

\usepackage{subfigure}
\usepackage{capt-of}
\usepackage{makecell}

\usepackage{mathtools}
\usepackage{pifont}

\title{\LARGE \bf
Multi-Granular Transformer for Motion Prediction with LiDAR
}

\author{Yiqian Gan$^*$,
        {Hao Xiao$^*$},
        {Yizhe Zhao$^*$},       
        {Ethan Zhang}, 
        {Zhe Huang},
        {Xin Ye},
        {Lingting Ge} \\
        TuSimple, Inc.
\thanks{ $*$ equal contribution
        }
}
\begin{document}

\maketitle
\thispagestyle{empty}
\pagestyle{empty}

%%%%%%%%%%%%%%%%%%%%%%%%%%%%%%%%%%%%%%%%%%%%%%%%%%%%%%%%%%%%%%%%%%%%%%%%%%%%%%%%
\begin{abstract}
Motion prediction has been an essential component of autonomous driving systems since it handles highly uncertain and complex scenarios involving moving agents of different types. In this paper, we propose a Multi-Granular TRansformer (MGTR) framework, an encoder-decoder network that exploits context features in different granularities for different kinds of traffic agents. To further enhance MGTR's capabilities, we leverage LiDAR point cloud data by incorporating LiDAR semantic features from an off-the-shelf LiDAR feature extractor. We evaluate MGTR on Waymo Open Dataset motion prediction benchmark and show that the proposed method achieved state-of-the-art performance, ranking 1st on its leaderboard~\footnote{https://waymo.com/open/challenges/2023/motion-prediction/}. 

\end{abstract}

\section{Introduction}
\label{sec:intro}
High-quality motion prediction in a long horizon is essential for the development of safety-critical autonomous vehicles. It serves as one of the cornerstones of related fields including scene understanding and decision-making in the realm of autonomous driving. Although advancements were made in past years, major challenges still exist and come from the following aspects: 
(i) Heterogeneous data acquired by autonomous vehicles such as maps and agent history states is non-trivial to be represented in a unified space. (ii) Environment context inputs from upstream modules including object detection and pre-built maps have limitations (e.g., uncountable amorphous regions such as bushes, walls, and construction zones, can be missing). (iii) Multimodal nature of agent behaviors brings further complexity. Here, the multimodal agent behaviors refer to discrete and diverse agent intents and possible futures. This work addresses these challenges with a proposed multi-granular Transformer model, namely MGTR.

\begin{figure}[t!]
        \centering
        \hspace{-3mm}
        \includegraphics[width=0.9\linewidth]{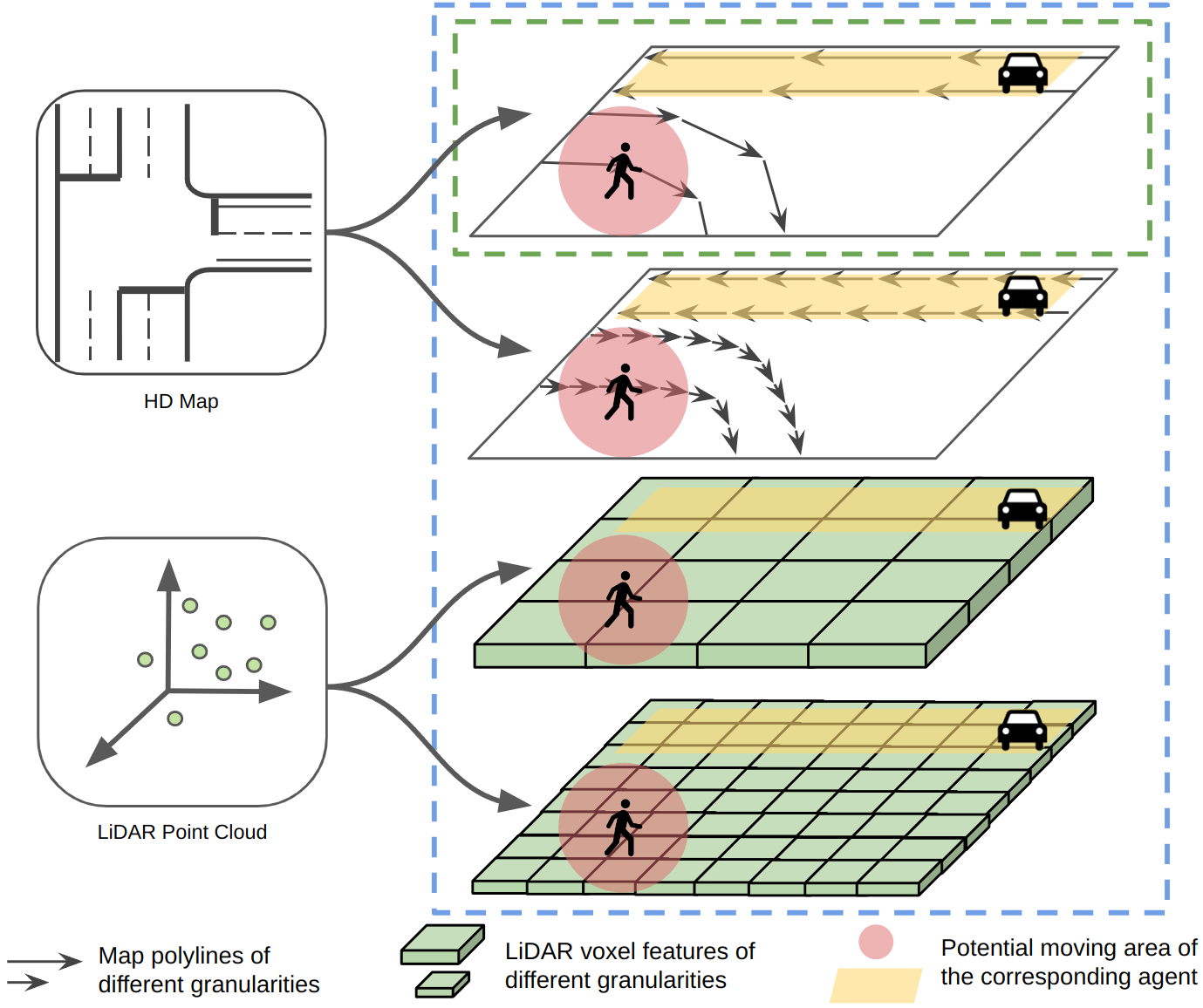}\\
        \vspace{-2mm}
        \caption{\textbf{Comparing context information used in different motion prediction frameworks.} Most previous methods \cite{varadarajan2022multipath++, shi2022motion} encode road graph only in a single granularity for all agents in the scene ({\color{ForestGreen}\textbf{green}} dashed box). In our method, various agents can benefit from multi-granular context information encoded from multimodal sources ({\color{RoyalBlue}\textbf{blue}} dashed box).}
        \label{fig:gtr_simple}
        \vspace{-7mm}
\end{figure}

Early methods mainly render inputs including High Definition (HD) map, agent history states into rasterized images, and apply convolutional neural networks (CNN) to encode scene information \cite{fang2020tpnet, wu2020motionnet, cui2019multimodal}. While convenient, long-range interactions are hard to capture in rasterization-based methods due to the limited receptive field of convolutions. A majority of recent works represent inputs as vectors by pre-processing raw continuous inputs into discrete sample points at a fixed sample rate~\cite{gao2020vectornet, varadarajan2022multipath++, shi2022motion}. Such vectors are later sent to graph-structured models for scene-level information extraction. A low sample rate will lose geometric details like curb curvature. On the other end, a high sample rate requires models' stronger ability to learn complex topographic relationships like lane centerline connectivity, which constrains perception range due to an increased number of nodes with limited computational resources. As shown in Fig. \ref{fig:gtr_simple}, in reality, different types of agents with varying motion patterns would benefit from context information at multiple distinguished granularities. 
Concurrently, the computer vision community has developed multiple ways of multi-granularity processing, such as feature pyramid techniques \cite{lin2017feature}. Through the years, it has been proven to be effective in gaining more comprehensive context information. 

Furthermore, the majority of existing motion prediction works are developed based on open datasets~\cite{ettinger2021large, Argoverse2} which usually provide 2D agent tracking states and static HD maps. However, in real world scenarios, a lot of other context information such as bushes, building walls, and traffic cones can also serve as strong cues for motion prediction. The new dataset WOMD-LiDAR\cite{chen2023womdlidar} enables us to incorporate LiDAR point cloud containing dense 3D context information for motion prediction. To the best of our knowledge, only a few works have combined LiDAR data into motion prediction. They primarily focus on extracting instance-level information \cite{li2023pedestrian, chen2023womdlidar} rather than rich environment context. Also, they do not explore LiDAR information in a multi-granular manner.

In this work, we propose a multi-granular Transformer model (MGTR), for motion prediction of heterogeneous traffic agents. MGTR follows a Transformer encoder-decoder architecture. It fuses multimodal inputs including agent history states, map elements, and extra 3D context embeddings from LiDAR. Both map elements and LiDAR embeddings are processed into sets of tokens at several granular levels for better context learning. Next, agent embeddings and multi-granular context embeddings are passed through a Transformer encoder after being filtered by our motion-aware context search for better efficiency. Then, motion predictions are generated through iterative refinement within the decoder and modeled by Gaussian Mixture Model (GMM). Our contributions can be summarized as follows: (i) We introduce a novel Transformer-based motion prediction method utilizing multimodal and multi-granular inputs, with motion-aware context search mechanism to enhance accuracy and efficiency. (ii) We present an approach to incorporate LiDAR inputs practically and efficiently for the purpose of motion prediction. (iii) We demonstrate state-of-the-art performance on Waymo Open Dataset motion prediction benchmark.

\section{Related work}
\textit{\textbf{Motion prediction}}: 
Early works \cite{djuric2020uncertainty, fang2020tpnet, cui2019multimodal, wu2020motionnet, chai2019multipath, marchetti2021mantra, park2020diverse, casas2019spatiallyaware, casas2021mp3} on motion prediction usually represent inputs as rasterized images, and adopts CNN to obtain high-quality results. For more direct context representation, VectorNet \cite{gao2020vectornet} brings up the vector representation that sequentially samples and connects map and agent history states into polylines, and combines them with graph-based models. 
Vectorized inputs enable researchers to tackle both structured and unstructured data, and build more versatile models \cite{zhao2020tnt, varadarajan2022multipath++, gu2021densetnt}. An emerging trend rises regarding Transformer-based models in various NLP and vision applications \cite{vaswani2023attention}. SceneTransformer \cite{ngiam2021scene} and Wayformer\cite{nayakanti2023wayformer}, combine the vector representation and apply a Transformer-based model to handle multimodal input. MTR\cite{shi2022motion} and MTR++ \cite{shi2023mtr} further improve vectorized representation by using local connected graphs and apply Transformer structures with their inspirations from DETR \cite{carion2020endtoend} and DAB-DETR \cite{liu2022dab}. In this work, we adopt the vectorized representation but introduce the multi-granular structure for multimodal input, which is an essential aspect neglected by previous works.

\textit{\textbf{Multi-granularity learning}}: 
The concept of multi-granularity learning originates from the field of computer vision. It is capable of learning patterns of various granular features \cite{lin2017feature, szegedy2015going, xiao2018group, lin2019group, fan2021multiscale}.
InceptionNet \cite{szegedy2015going} and FPN \cite{lin2017feature} have shown great success in image classification tasks by incorporating multi-granularity techniques into CNN models, while MViT \cite{fan2021multiscale} proves the effectiveness of multi-granularity representation in Transformer-based models. In this work, we exploit advantages of applying multi-granularity to Transformer-based models for motion prediction, which has rarely been explored. 

\textit{\textbf{LiDAR for motion prediction}}: 
Applying LiDAR in the field of motion prediction is fairly new. In recent years, with the advancement of multimodal learning, researchers have been trying to incorporate LiDAR data towards different learning tasks in the field \cite{casas2018intentnet, djuric2021multixnet,laddha2021mvfusenet, li2023pedestrian, chen2023womdlidar}. Most work such as IntentNet \cite{casas2018intentnet} and MultiXNet \cite{djuric2021multixnet} take LiDAR data as the only perception input and generate both object detection and motion prediction through multi-task settings. More Recent works \cite{li2023pedestrian, chen2023womdlidar} focus on motion prediction and mainly use LiDAR as instance-level features. Unlike previous work, the proposed MGTR treats LiDAR data as context features and incorporates the advantages of multimodal learning.

% \vspace{-5mm}
\begin{figure*}
    \centering
    \includegraphics[scale=0.23]{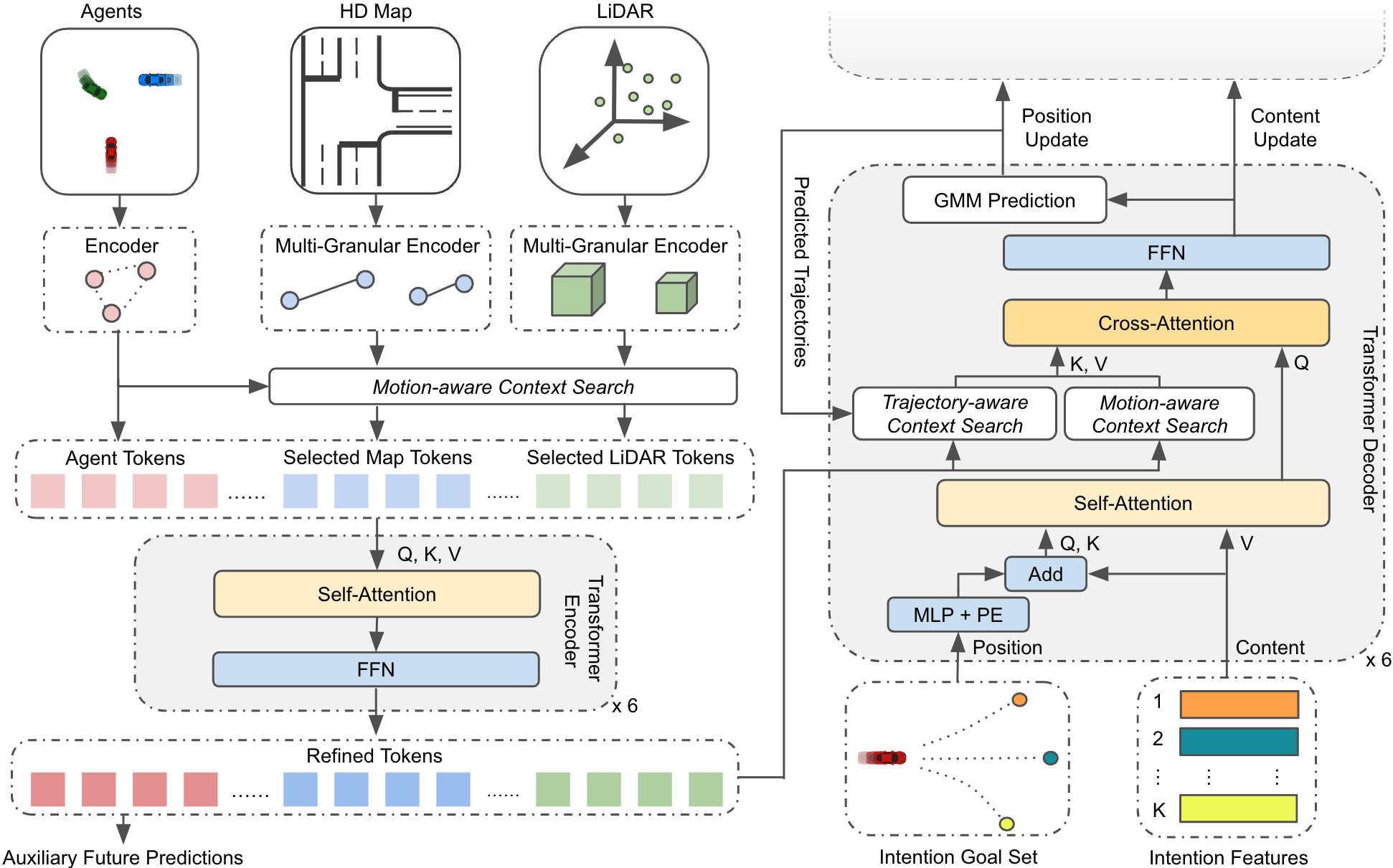}
    % \vspace{-3mm}
    \caption{\textbf{An overview of our proposed MGTR.} Agent trajectories and map elements are represented as polylines and encoded as agent and multi-granular map tokens. LiDAR data is processed by a pre-trained model into voxel features and further transformed into multi-granular LiDAR tokens. Motion-aware context search selects a set of map and LiDAR tokens, refined together with agent tokens through local self-attention in the Transformer encoder. Finally, a set of intention goals and their corresponding content features are sent to the decoder to aggregate context features. Multiple future trajectories of each agent will be predicted based on its intention goals, supporting the multimodal nature of agent behaviors.}
    \label{fig:architecture}
    \vspace{-5mm}
\end{figure*}

\section{Method}

As depicted in Fig. \ref{fig:architecture}, we proposed the MGTR model, a novel Transformer-based framework that takes multimodal inputs in a multi-granular manner including LiDAR data. In \ref{sec:input_representation}, we first introduce how different inputs are represented and encoded into multi-granular tokens and how the number of tokens is reduced by motion-aware context search. Next, in \ref{sec:gtr_encoder} and \ref{sec:gtr_decoder}, we demonstrate how tokens are refined in the encoder and utilized in the decoder for motion prediction. Finally, \ref{sec:loss} introduces the training losses used in our model. 

\subsection{Multimodal Multi-Granular Inputs}\label{sec:input_representation}

\subsubsection{\textbf{Agent and map}}
Following representation in VectorNet \cite{gao2020vectornet}, agent state history is sampled at a constant time interval and processed into vectorized polylines as $\text{P}_\text{A} \in \mathbb{R}^{N_a \times T_h \times C_a}$ to represent state information from $T_0 - T_h$ to $T_0$, where $N_a$ denotes the number of target agents in a scene, $C_a$ as the dimension of agent features, $T_0$ as the current time, $T_h$ as the time horizon. The agent state features $C_a$ include position, velocity, 3D bounding box size, heading angle, object type, etc. Zero paddings are added in the time dimension if the tracking length is smaller than $T_h$. Then, each agent polyline will first be transformed into the target agent-centric coordinate followed by a PointNet-like~\cite{qi2017pointnet} polyline encoder as shown in Eq. \ref{eq:agent_map_polyline}.

 Different types of agents have different movement ranges and requirements for map granularity. In this work we extract map contents in a multi-granular manner. Map elements with topological relationships such as road centerlines and area boundaries are sampled evenly at different sample rates, resulting in polylines with different granularities. Concretely, we represent sets of multi-granular polylines as $\{ \text{P}_\text{M}^{(i)} \} \in \mathbb{R}^{N_m^{(i)}\times N_s^{(i)} \times C_m}$, where $\text{P}_\text{M}^{(i)}$ denotes map polylines at $i$-th spatial granularity, $N_m^{(i)}$ denotes the number of polylines at $i$-th granularity, $N_s^{(i)}$ denotes the number of sampled points in each polyline, $C_m$ denotes the token feature dimension for map including positions, curvature, speed limit, etc. With a high sample rate, the same road centerline will be sampled into more polylines similar to more image pixels on high-resolution images. $N_{m}^{(i)}$ polylines are generated at each sample rate $r_i$. Similar to agent polylines, each map polyline is transformed to an agent-centric coordinate and encoded by a PointNet-like structure as:
 
\vspace{-2mm}
\begin{align}\label{eq:agent_map_polyline}
F_{\text{A}}       = \phi\Big(\text{MLP}\big(\Gamma ( \text{P}_\text{A}) \big) \Big),~~
F_{\text{M}}^{(i)} = \phi\Big(\text{MLP}\big(\Gamma ( \text{P}_\text{M}^{(i)} \big) \Big),
\end{align}
where $\Gamma (\cdot)$ denotes the coordinate transformation, $\text{MLP}(\cdot)$ denotes a multi-layer perceptron, $\phi$ denotes max-pooling. Agent and map polylines are encoded into $F_{\text{A}} \in \mathbb{R}^{N_a\times C}$ and $F_{\text{M}}^{(i)} \in \mathbb{R}^{N_{m}^{(i)}\times C}$ with a feature dimension of $C$. The weights of the polyline encoders for different granularities are not shared, to ensure map features at each granularity are kept.

\subsubsection{\textbf{LiDAR}} \label{method-lidar}
In order to obtain richer 3D context information missing in explicit perception outputs and pre-built HD maps, we propose to integrate LiDAR information into our framework. Using raw LiDAR point cloud directly in the motion prediction network is inefficient and resource-intensive, due to its sparsity and large magnitude. Therefore, we choose to use LiDAR voxel features extracted by an off-the-shelf LiDAR segmentation network. This typically does not add additional overhead to autonomous driving systems since most deployed systems have such network already implemented. 
Although MGTR is not restricted to a certain LiDAR module, we adopt a voxel-based segmentation network, specifically LidarMultiNet \cite{ye2023lidarmultinet}, as our pre-trained LiDAR model. We extract a voxel feature map $\mathcal{V}_{raw} \in \mathbb{R}^{C_{l}\times D\times H\times W}$ from an intermediate layer, where $C_{l}$ denotes the feature dimension, $D, H, W$ are the sizes of the voxel space.
It serves as a perfect input feature to represent context information for motion prediction. To add more context information, we concatenate one-hot embedding of the predicted semantic label of each voxel from the segmentation result and the 3D position of the center of each voxel with $C_{l}$ LiDAR segmentation features. The voxel feature becomes $\mathcal{V} \in \mathbb{R}^{C_{v}\times D\times H\times W}$, where $C_{v}$ is the LiDAR feature dimension after concatenation. 

To obtain LiDAR context features in multi-granularities and reduce complexity, we employ average pooling across various scales to obtain features of different granularities. 
Pooled features are encoded into tokens through an MLP as: 
\vspace{-1mm}
\begin{align}\label{eq:lidar_token}
F_{\text{L}}^{(i)} = \text{MLP}\Big( \mathcal{P}^{(i)} \big( \Gamma (\mathcal{V}) \big) \Big), 
\end{align}
where $\Gamma (\cdot)$ denotes the coordinate transformation, $\mathcal{P}^{(i)} (\cdot)$ denotes the average pooling for the $i$-th granularity. LiDAR features are encoded into $F_{\text{L}}^{(i)} \in \mathbb{R}^{N_{l}^{(i)}\times C}$, where $N_{l}^{(i)}$ is the number of LiDAR tokens for the $i$-th granularity, and $C$ is the feature dimension.

\subsubsection{\textbf{Motion-aware context search}}\label{sec:motion_aware} 
After multi-granular encoders, the number of raw tokens $N = N_a + \sum_{i} N_{m}^{(i)} + \sum_{i} N_l^{(i)}$ can be extremely large, making it impossible to send them directly to Transformer encoder due to computing resource constraint. To learn features more efficiently, we introduce motion-aware context search which helps boost training efficiency and encode more meaningful context for agents with different motion patterns. For agents with different velocities, the desired positions of scene context for long-horizon trajectory prediction differ significantly. Therefore, for an agent of interest, we use its current velocity to project a future distance as the context token search prior. Through the projected position, we acquire ${\widetilde{N}_m}$ nearest map tokens and ${\widetilde{N}_l}$ nearest LiDAR tokens, resulting in a total of $N_a + {\widetilde{N}_m} + {\widetilde{N}_l}$ selected tokens that will be fed into our Transformer encoder for further refinement.

\subsection{Transformer Encoder}\label{sec:gtr_encoder}

% 1. token inputs learning
\subsubsection{\textbf{Token aggregation and encoding}}
After the aforementioned vectorization and token generation, a Transformer encoder is established to aggregate features from multi-granular tokens. All tokens are refined through layers of encoder structure with a self-attention layer followed by a feed-forward network (FFN). To boost training efficiency and better capture neighboring information, a local attention mechanism is adopted. Let $F_e^{j}$ be the refined tokens output by the $j$-th layer. The multi-head self-attention\cite{vaswani2017attention} can be formulated as:
% \hspace{-5mm}
\begin{equation} \label{eq:mha}
\begin{aligned}
    Q &= F_e^{j-1} + PE(F_e^{j-1}),\quad V = \kappa(F_e^{j-1}), \\
    K &= \kappa(F_e^{j-1}) + PE(\kappa(F_e^{j-1})), \\
    F_e^j &= \text{MHSA}(Q, K, V),
\end{aligned}
\end{equation}
where $\kappa(\cdot)$ is a function that returns k-nearest neighboring tokens for each query token, $PE(\cdot)$ is a positional encoding function. $\text{MHSA}(\cdot,\cdot,\cdot)$ stands for multi-head self-attention layer \cite{vaswani2017attention}.

\subsubsection{\textbf{Future state enhancement}} \label{sec:aux_task}
In addition to considering agents' history trajectories, we also take into account their potential future trajectories, which play a crucial role in predicting the motion of agent of interest. Therefore, after agent tokens are refined by the Transformer encoder, a future trajectory is predicted for each agent following \cite{shi2022motion} and it can be formulated as:
\begin{equation} \label{eq:classification}
    \mathcal{T}_{scene} = \text{MLP}(F_e^A),
\end{equation}
where $\mathcal{T}_{scene} \in \mathbb{R}^{N_a \times T \times 4}$ denotes trajectories (including position and velocity) of $N_a$ agent for future $T$ frames and $F_e^A$ is the agent token from the encoder.
The future trajectories are further encoded by a polyline encoder and fused with the original agent token $F_e^{A}$ to form a future-aware agent feature that is fed into the decoder later. It's worth noting that $\mathcal{T}_{scene}$ is supervised by the ground truth trajectories, resulting in an auxiliary loss which will be introduced in \ref{sec:loss}. 

\subsection{Transformer Decoder}\label{sec:gtr_decoder}
\subsubsection{\textbf{Intention goal set}}
We generate $\mathcal{K}$ representative intention goals by adopting K-means clustering algorithm on endpoints of ground truth trajectories for different types of agents. Each intention goal represents an implicit motion mode, which can be modeled as a learnable positional embedding.

\subsubsection{\textbf{Token aggregation with intention goal set}}
In each layer, we first apply the self-attention module to propagate information among $\mathcal{K}$ intention queries as follows:
\hspace{-2mm}
\begin{equation} \label{eq:decoder_mha}
\begin{aligned}
    Q = K &= F_d^{j-1} + PE(F_d^{j-1}),\quad V = F_d^{j-1}, \\
    F_I^j &= \text{MHSA}(Q, K, V),
\end{aligned}
\end{equation}
where $F_d^{j-1} \in\mathbb{R}^{\mathcal{K}\times C_{dec}}$ are intention features from $(j\text{-}1)$-th decoder layer and $F_I^j$ is the updated intention feature, where $\mathcal{K}$ denotes number of intention goals, $C_{dec}$ denotes feature dimension. We initialize the intention features $F_d^0$ to be all zeros as the input for the first Transformer decoder. Next, a cross-attention layer is adopted for aggregating features from the encoder as:

\hspace{-5mm}
\begin{equation} \label{eq:decoder_mhca}
\begin{aligned}
    Q &= F_I^{j} + PE(F_I^{j}), \\
    K = V &= \gamma(F_e) + PE(\gamma(F_e)), \\
    F_d^j &= \text{MHCA}(Q, K, V),
\end{aligned}
\end{equation}
\hspace{-5mm}
\begin{equation} \label{eq:decoder_mhca}
    \gamma(F_e) = \eta(F_e) \cup \theta(F_e),
\end{equation}
where $F_I^j$ is the updated intention feature from previous multi-head self-attention. $F_e$ is the multi-granular context tokens from encoder, which includes future-aware agent tokens, map tokens, and LiDAR tokens. $\text{MHCA}(\cdot,\cdot,\cdot)$ is the multi-head cross-attention layer \cite{vaswani2017attention}. $\gamma(\cdot)$ is the combination of our trajectory-aware context search ($\eta(\cdot)$) and motion-aware context search ($\theta(\cdot)$), which intends to extract multi-granular context features from a local region. Inspired by the dynamic map collection from \cite{shi2022motion}, we introduce trajectory-aware context search which takes the predicted trajectories from previous decoder layer and selects the context token whose centers are close to the predicted trajectory. Along with the previously mentioned motion-aware context search, it continuously attends to the most important context information throughout iterative prediction refinement.

\subsubsection{\textbf{Multimodal motion prediction with GMM}}
Following \cite{varadarajan2022multipath++, shi2022motion}, we model the multimodal future trajectories with GMM. 
For each decoder layer, we append a classification head and a regression head for the intention feature $F_d^j$ respectively:
\begin{equation} \label{eq:classification}
    p = \text{MLP}(F_d^j), \quad  \mathcal{T}_{target} = \text{MLP}(F_d^j),
\end{equation}
where $p \in \mathbb{R}^{\mathcal{K}}$ is the probability distribution of each trajectory mode corresponding to each intention goal.
$\mathcal{T}_{target} \in \mathbb{R}^{\mathcal{K} \times T \times 7}$ is predicted GMM parameters representing $\mathcal{K}$ future trajectories and 2D velocities for $T$ future frames. 
The endpoints of predicted trajectories will be used for positional embedding in the next decoder layer.

\subsection{Training Loss}\label{sec:loss}
The training loss in this work is a weighted combination of: (i) auxiliary task loss $\mathcal{L}_{aux}$ on future predicted trajectories of all agents $\mathcal{T}_{scene}$ (ii) classification loss $\mathcal{L}_{cls}$ in form of cross entropy loss on predicted intention probability $p$ (iii) GMM loss $\mathcal{L}_{GMM}$ in form of negative log-likelihood loss of the predicted trajectories of target agent $\mathcal{T}_{target}$.  Auxiliary task loss is measured with L1 loss between ground truth and predictions of both agents' position and velocity. Similar to \cite{shi2022motion}, we use a hard-assignment strategy that selects the best matching mode and calculates the GMM loss and the classification loss. This can force each mode to specialize for a distinct agent behavior.

\section{Experiments}

\subsection{Dataset}

Following most recent motion prediction works~\cite{shi2022motion, chen2023womdlidar, li2023pedestrian}, we conduct extensive experiments on WOMD-LiDAR dataset~\cite{chen2023womdlidar}. It is a large-scale motion dataset with LiDAR specifically designed for motion prediction in the field of autonomous driving, containing 100,000+ real-world driving video clips with diverse driving scenarios (e.g. intersection, lane merging). It shares the same training, evaluation and test samples with its predecessor WOMD~\cite{ettinger2021womd}. WOMD-LiDAR contains 3 categories of agents of interest: vehicle, pedestrian, and cyclist. 
The performance of this task on WOMD-LiDAR is measured by the minimum Average Displacement Error (minADE), the minimum Final Displacement Error (minFDE), miss rate (MR) and the mean Average Precision (mAP) of predicted trajectories in the next 3, 5 and 8 seconds, with mAP as the major evaluation metric. Details of those metric definitions can be found in~\cite{ettinger2021womd}.

\subsection{Experiment Setup}

\subsubsection{\textbf{LiDAR voxel encoder}}

We use LidarMultiNet~\cite{ye2023lidarmultinet} as our LiDAR encoder to extract LiDAR voxel features from raw point cloud. Resulting 3D voxel features after its global context pooling are used as our prediction model input. The original feature dimension $C_l$ is 32. As discussed in Sec~\ref{method-lidar}, for each voxel, we concatenate a 22-dim predicted one-hot segmentation result as well as its 3-dim position by channel, formulating our 57-dim LiDAR voxel features (i.e. $C_v = 57$). Average pooling is applied twice on original voxels to form multi-granularities, making the length and width of each LiDAR voxel 0.8m or 1.6m. LidarMultiNet is pre-trained on Waymo Open Dataset~\cite{sun2020swod} and is frozen during the training of our model. 

\subsubsection{\textbf{Implementation details}}

For full-scale experiments on WOMD-LiDAR \texttt{val} set and \texttt{test} set, our batch size is 10 per GPU and we train from scratch for 30 epochs. We use AdamW optimizer~\cite{loshchilov2017adamw} with an initial learning rate of 0.0001, in conjunction with a multi-step scheduler. The learning rate is decayed by a factor of 0.5 every two epochs after 20 epochs. Both the encoder and the decoder consist of 6 Transformer layers. We adopt ${\widetilde{N}_m} = 768$ map polylines as the topographic context in the encoder. In addition, we add ${\widetilde{N}_l} = 256$ multi-granular LiDAR voxels to complement our local scene representation.
In practice, the length of each map polyline is either 10 map points or 20 map points, equivalent to 5m or 10m in range. The number of neighbors in local self-attention of the encoder is set to 32.
For each sample, we predict 64 candidate trajectories using 64 intention goals, which are generated from K-means clustering over all ground truth goal points (at 8-second prediction horizon) in the training set.
Non-maximum suppression (NMS) is applied to post-process predictions, resulting in 6 final trajectories per sample.

For ablation study, unless stated otherwise, all experiments use a same hyperparameter set as the full-scale experiments. We only use 20\% of the total training data in WOMD-LiDAR via a fixed sampler for efficiency purposes.

\begin{figure*}
        \centering
        \includegraphics[width=0.9\linewidth]{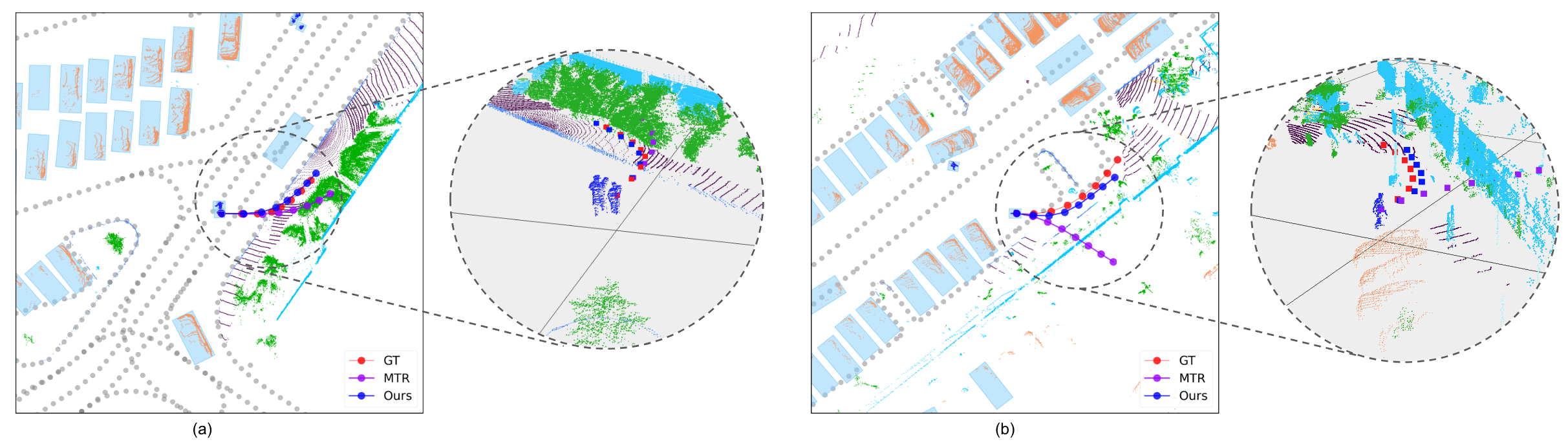}
        \caption{\textbf{Visualization of prediction result comparison between MTR \cite{shi2023mtr} and MGTR (Ours)}. A global bird's-eye-view (including agents, HD map and LiDAR point cloud) and a local LiDAR visualization for each scene. For LiDAR point cloud, only limited semantic class such as vegetation ({\color{Green}\textbf{green}} points), building ({\color{Cyan}\textbf{cyan}} points), sidewalk ({\color{Brown}\textbf{brown}} points), vehicle({\color{Orange}\textbf{orange}} points) and pedestrian ({\color{Blue}\textbf{blue}} points) are shown for better visualization.}
        \label{fig:qual}
\end{figure*}

\subsection{Quantitative Analysis}

\begin{table}
\centering
\caption{\textbf{Comparison on WOMD-LiDAR \texttt{val} set.} Results in top three rows are computed as an average of $t=3, 5$, and $8$ seconds, while the ones in bottom three rows are reported for $t=8$ seconds. MTR++ \cite{shi2023mtr} does not report categorical results. Following~\cite{chen2023womdlidar}, all metrics are reported with two decimal places. $\ast$ indicates methods utilizing LiDAR.}
\label{tab:main_val_set}
\tiny
\resizebox{0.48\textwidth}{!}{
\setlength{\tabcolsep}{3.5pt}
% @{\vline}
\begin{tabular}{
l@{\hspace{1.0\tabcolsep}}
c@{\hspace{1.0\tabcolsep}}
c@{\hspace{1.0\tabcolsep}}
c@{\hspace{1.0\tabcolsep}}
c}

\toprule
 & \multicolumn{4}{c}{\textbf{mAP} $\uparrow$} \\
 \cmidrule(lr){2-5}
\textbf{Method} & Vehicle & Pedestrian &Cyclist & Average \\
\midrule
MTR~\cite{shi2022motion} & 0.45 & 0.44 & 0.36 & 0.42 \\
MTR++~\cite{shi2023mtr} & - & - & - & 0.44 \\
\textbf{MGTR}${}^\ast$ (Ours) & \textbf{0.46} & \textbf{0.47} & \textbf{0.40} & \textbf{0.45} \\
\midrule
% LSTM~\cite{ettinger2021womd}                        & 0.23 & 0.23 & 0.21 & 0.22 \\
% SceneTransformer~\cite{ngiam2021scene} & 0.26 & 0.27 & 0.20 & 0.24 \\
Wayformer~\cite{nayakanti2023wayformer}             & 0.35 & 0.35 & 0.29 & 0.33 \\
Wayformer+LiDAR~\cite{chen2023womdlidar}${}^\ast$   & 0.37 & 0.37 & 0.28 & 0.34 \\
\textbf{MGTR}${}^\ast$ (Ours)                      & \textbf{0.38} & \textbf{0.44} & \textbf{0.32} & \textbf{0.38} \\
\bottomrule
\end{tabular}}
\vspace{-0.2cm}
\end{table}

\begin{table*}
  \centering
  \caption{\textbf{Comparison on WOMD-LiDAR \texttt{test} set}. All metrics are averaged over 3s, 5s, and 8s. All models do not use model ensemble.}
  \label{tab:main_test_set}
  \tiny
  \resizebox{0.98\textwidth}{!}{
  \setlength{\tabcolsep}{3.5pt}
  \begin{tabular}{l
                  c@{\hspace{1.0\tabcolsep}}c@{\hspace{1.0\tabcolsep}}c@{\hspace{1.0\tabcolsep}}c
                  c@{\hspace{1.0\tabcolsep}}c@{\hspace{1.0\tabcolsep}}c@{\hspace{1.0\tabcolsep}}c
                  c@{\hspace{1.0\tabcolsep}}c@{\hspace{1.0\tabcolsep}}c@{\hspace{1.0\tabcolsep}}c
                  c}
    
    \toprule
    & \multicolumn{4}{c}{Vehicle} & \multicolumn{4}{c}{Pedestrian} & \multicolumn{4}{c}{Cyclist} & \multicolumn{1}{c}{Avg} \\
    \cmidrule(lr){2-5} \cmidrule(lr){6-9} \cmidrule(lr){10-13} \cmidrule(lr){14-14}
    \textbf{Method} & \textbf{minADE}$\downarrow$ & \textbf{minFDE}$\downarrow$  & \textbf{MR}$\downarrow$ & \textbf{mAP}$\uparrow$
                    & \textbf{minADE}$\downarrow$ & \textbf{minFDE}$\downarrow$  & \textbf{MR}$\downarrow$ & \textbf{mAP}$\uparrow$
                    & \textbf{minADE}$\downarrow$ & \textbf{minFDE}$\downarrow$  & \textbf{MR}$\downarrow$ & \textbf{mAP}$\uparrow$
                    & \textbf{mAP}$\uparrow$ \\
    \midrule
    ReCoAt~\cite{huang2022recoat} & 0.9865 & 2.1771 & 0.2695 & 0.2667 & 0.4261 & 0.8982 & 0.1451 & 0.3208 & 0.8985 & 1.9252 & 0.3164 & 0.2258 & 0.2711 \\
    DenseTNT~\cite{gu2021densetnt} & 1.3462 & 1.9120 & 0.1518 & 0.3698 & 0.5013 & 0.9130 & 0.1014 & 0.3342 & 1.2687 & 1.8292 & 0.2186 & 0.2802 & 0.3281 \\
    SceneTransformer~\cite{ngiam2021scene} & \textbf{0.7094} & \textbf{1.4115} & 0.1480 & 0.3270 & 0.3812 & 0.7532 & 0.0971 & 0.2715 & 0.7446 & 1.4701 & 0.2239 & 0.2380 & 0.2788 \\ % leaderboard from whchat
    GTR-R36~\cite{liu2023gtrr36} & 0.7450 & 1.5049 & 0.1477 & 0.4521 & 0.3470 & 0.7221 & 0.0741 & 0.4243 & 0.7095 & 1.4406 & 0.1772 & 0.4003 & 0.4255 \\ % leaderboard GTR-R36
    DM~\cite{yu2023dm} & 0.7701 & 1.5400 & 0.1529 & 0.4725 & 0.3741 & 0.7882 & 0.0848 & 0.4172 & 0.7436 & 1.4885 & 0.2043 & 0.4005 & 0.4301 \\ % leaderboard DM
    MTR~\cite{shi2022motion} & 0.7642 & 1.5257 & 0.1514 & 0.4494 & 0.3486 & 0.7270 & 0.0753 & 0.4331 & 0.7022 & \textbf{1.4093} & 0.1786 & 0.3561 & 0.4129 \\ % MTR Tab 5
    MTR++~\cite{shi2023mtr} & 0.7178 & 1.4321 & \textbf{0.1366} & \textbf{0.4871} & 0.3504 & 0.7305 & 0.0745 & 0.4324 & 0.7036 & 1.4190 & 0.1784 & 0.3792 & 0.4329 \\ % leaderboard mtr++
    \textbf{MGTR} (Ours) & 0.7393 & 1.5119 & 0.1497 & 0.4626 & \textbf{0.3441} & \textbf{0.7191} & \textbf{0.0722} & \textbf{0.4865} & \textbf{0.6919} & 1.4096 & \textbf{0.1675} & \textbf{0.4023} & \textbf{0.4505} \\
    \bottomrule
  \end{tabular}}
  \vspace{-0.2cm}
\end{table*}

We report quantitative results on WOMD-LiDAR \texttt{val} set, as shown in TABLE~\ref{tab:main_val_set}. We compare MGTR against other state-of-the-art models. Remarkably, we achieve the best mAP over all types of agents. Specifically, compared to Wayformer+LiDAR~\cite{chen2023womdlidar}, a multimodal model with LiDAR input, our model substantially improves mAP on pedestrians by 7\%, cyclists by 4\%, and vehicle by 1\% for $t=8$ seconds. MGTR demonstrates similar advancement compared with previous SOTA MTR \cite{shi2022motion} and MTR++ \cite{shi2023mtr}. We argue that this indicates MGTR is better at capturing subtle movements thanks to the multi-granular representation of both map and LiDAR. 

Furthermore, we achieve the state-of-the-art performance on WOMD-LiDAR \texttt{test} set. As revealed in TABLE ~\ref{tab:main_test_set}, the single-model version of our proposed MGTR achieves an overall mAP of 45.05\%, which has +1.76\% advantage over the second-best model, significantly outperforming all other single-model entries. Compared to the latest state-of-the-art motion prediction model, MTR++~\cite{shi2023mtr}, we achieve a whopping 5.41\% increase in terms of mAP on the pedestrian category. Apart from mAP, our model also improves over a variety of metrics, including minADE, minFDE, MR for multiple categories. Specifically, we reduce trajectory miss rate (MR) on cyclists by 1.11\% in comparison with the second-best model. This strongly signals that for non-vehicular objects, features that attend to details are key to more accurate and reliable trajectory predictions.

As of the paper submission date (Sep. 15, 2023), it is worth noting that our proposed MGTR with model ensemble has ranked \textbf{first place} among all submissions on the motion prediction track of Waymo Open Challenge Leaderboard~\footnote{https://waymo.com/open/challenges/2023/motion-prediction/}. Detailed comparison of methods with model ensemble techniques is not included in the paper due to page limits.

\subsection{Qualitative Analysis}
We further delve into the qualitative aspects of our proposed MGTR model using visualizations. 
While our model is capable of generating multimodal trajectories, for the sake of illustration, we have chosen to visualize the trajectory with the highest probability, demonstrating the efficacy of our approach.
Fig. \ref{fig:qual} showcases two scenarios in which a pedestrian is crossing a street. When we examine the trajectory predicted by MTR \cite{shi2023mtr}, in the scenarios the pedestrian will (a) walk into the bushes and (b) pass through a building, which are not reasonable outcomes. In contrast, our MGTR model successfully predicts that the pedestrian will walk onto the sidewalk, skillfully (a) avoiding any collision with the bushes, and (b) avoiding the building. 
These visualizations underscore the superior predictive capabilities of MGTR in comparison to the MTR, particularly in situations where LiDAR can provide additional 3D context information that is not available within HD map.

\subsection{Ablation Study}

We conduct ablation studies on WOMD-LiDAR $val$ set, as shown in TABLE ~\ref{tab:abla}. The baseline has a similar structure as MGTR, but only possesses single coarse-granular map tokens. By adding our proposed designs one by one, we observe consistent and notable mAP improvements for all types of agents.

\begin{table}
\centering
\caption{\textbf{Ablation study on our proposed MGTR.}}
\label{tab:abla}
\tiny
\resizebox{0.48\textwidth}{!}{
\setlength{\tabcolsep}{3.5pt}
% @{\vline}
\begin{tabular}{
l@{\hspace{1.0\tabcolsep}}
c@{\hspace{1.0\tabcolsep}}
c@{\hspace{1.0\tabcolsep}}
c@{\hspace{1.0\tabcolsep}}
c}

\toprule
 & \multicolumn{4}{c}{\textbf{mAP} $\uparrow$} \\
 \cmidrule(lr){2-5}
\textbf{Description} & Vehicle & Pedestrian &Cyclist & Average \\
\midrule
Baseline & 0.3860 & 0.3682 & 0.2881 & 0.3474 \\
+ multi-granular map & 0.3895 & 0.3730 & 0.2900 & 0.3508 \\
+ multi-granular LiDAR & 0.3896 & 0.3820 & 0.2997 & 0.3571 \\
+ motion-aware context search & \textbf{0.3919} & \textbf{0.3935} & \textbf{0.3025} & \textbf{0.3626} \\
\bottomrule
\end{tabular}}
\vspace{-0.2cm}
\end{table}

\subsubsection{\textbf{Multi-granular map}}

The multi-granular design allows our model to represent the world at different resolutions so that agents can benefit from the granularity that suits them. Compared to the baseline, adding multi-granularity significantly improves mAP of all types, especially for pedestrians.

\subsubsection{\textbf{Multi-granular LiDAR}}
Voxelized LiDAR features contain fine-grained scene information that can affect agents' future behaviors but may not be covered by perception results and HD map. Adopting LiDAR as context features improves mAP of pedestrians and cyclists drastically.

\subsubsection{\textbf{Motion-aware context search}}

Compared to the original context search introduced in~\cite{shi2023mtr} that does not consider individual moving patterns, our motion-aware context search factors in motion compensation on an individual basis. After adding this feature, we observe significant improvements on mAP. It indicates that our design can better accommodate different agents of interest to retrieve relevant context information, thus improving motion prediction accuracy.

\section{Conclusion}

In this paper, we propose MGTR, a novel Transformer-based motion prediction model that incorporates multimodal inputs including LiDAR point cloud in an effective multi-granular manner. Rich context features at different granularities enhance the overall motion prediction performance. 
Our model reaches state-of-the-art performance on the public WOMD-LiDAR motion prediction benchmark with significant improvements over pedestrians and cyclists.

% \section*{Appendix}

%%%%%%%%%%%%%%%%%%%%%%%%%%%%%%%%%%%%%%%%%%%%%%%%%%%%%%%%%%%%%%%%%%%%%%%%%%%%%%%%

\bibliographystyle{IEEEtran}
\bibliography{reference}

\end{document}